\documentclass[fleqn,10pt]{wlscirep}
\usepackage[utf8]{inputenc}
\usepackage[T1]{fontenc}

\usepackage{hyperref}
\def\numberofsectionsinstudy{95 }
\def\numberofsubjectsinstudy{17 }
\def\numberofcellsinrcm{4,503 }
\def\etal{\textit{et al.} }
\usepackage{subcaption}
\usepackage{siunitx}
\usepackage{float}
\usepackage{caption}
\usepackage{graphicx}
\usepackage{array}     
\usepackage{ifthen}

\def\titletext{Leveraging Computational Pathology AI for Noninvasive Optical Imaging Analysis Without Retraining}
\title{\titletext}

\author[1]{Danny Barash}
\author[2]{Emilie Manning}
\author[2]{Aidan Van Vleck}
\author[1]{Omri Hirsch}
\author[2]{Kyi Lei Aye}
\author[3]{Jingxi Li}
\author[4]{Philip O. Scumpia}
\author[3]{Aydogan Ozcan}
\author[5]{Sumaira Aasi}
\author[6]{Kerri E. Rieger}
\author[5]{Kavita Y. Sarin}
\author[1]{Oren Freifeld}
\author[2,7,*]{Yonatan Winetraub}
\affil[1]{Department of Computer Science, Ben Gurion University, Be'er Sheva, 8499000, Israel.}
\affil[2]{Department of Structural Biology, Stanford University, Stanford, CA 94305, USA.}
\affil[3]{Electrical and Computer Engineering Department, University of California, Los Angeles, CA, 90095, USA.}
\affil[4]{Division of Dermatology, University of California, Los Angeles, CA, 90095, USA.}
\affil[5]{Department of Dermatology, Stanford University School of Medicine, Stanford, CA 94305, USA.}
\affil[6]{Department of Pathology, Stanford University
School of Medicine and Stanford Cancer Institute, Stanford, CA 94305, USA.}

\affil[7]{The Bio-X Program, Stanford, CA 94305, USA.}

\affil[*]{yonatanw1@stanford.edu}

\begin{abstract}
Noninvasive optical imaging modalities can probe patient’s tissue in 3D and over time generate gigabytes of clinically relevant data per sample. There is a need for AI models to analyze this data and assist clinical workflow. The lack of expert labelers and the large dataset required (>100,000 images) for model training and tuning are the main hurdles in creating foundation models. In this paper we introduce FoundationShift, a method to apply any AI model from computational pathology without retraining. We show our method is more accurate than state of the art models (SAM, MedSAM, SAM-Med2D, CellProfiler, Hover-Net, PLIP, UNI and ChatGPT), with multiple imaging modalities (OCT and RCM). This is achieved without the need for model retraining or fine-tuning. Applying our method to noninvasive in vivo images could enable physicians to readily incorporate optical imaging modalities into their clinical practice, providing real time tissue analysis and improving patient care. 
\end{abstract}

\begin{document}
\flushbottom
\maketitle
\thispagestyle{empty}

\ifthenelse{\boolean{false}}{ 
\begin{figure}[t!]
\centering
\includegraphics[width=1\linewidth]{images/Visual Abstract.png}
\caption*{\textbf{Visual Abstract}}
\label{fig:visualabstract}
\end{figure}
}{}

\section*{Main}

Noninvasive optical imaging modalities such as Optical Coherence Tomography (OCT), Reflectance Confocal Microscopy (RCM), photoacoustic microscopy and others have shown promise as noninvasive diagnostic tools~\cite{fisher2018clinical,ulrich2015sensitivity,li2014high} able to track disease and treatment over time, in 3D, and in high resolution~\cite{kuck2014evaluation, tsai2021submicron}. 
Every scan produces gigabytes of clinical data, posing a significant challenge for physicians to analyze it manually.

To support clinical decision-making, Artificial Intelligence (AI) models can be leveraged. However, the relatively small amounts of available data, due to a lack of clinical expert annotators and the diversity of imaging modalities~\cite{lin2024rapid,kumar2023deep}, have limited development to specialized machine learning models designed to solve specific tasks. There is a growing interest in transitioning from these narrow AI models to more generalizable foundation models.

Foundation models are AI models that can be adapted for a wide range of tasks. They are trained once on large annotated datasets (exceeding 100,000 images) and can then be applied to solve a multitude of problem classes. 

A key limitation of using foundation models in optical imaging analysis is the model's reliance on similar annotated data in the training sets, which is often scarce. Consequently, foundation models significantly underperform when evaluating OCT images compared to more established modalities, such as histopathology Hematoxylin and Eosin (H\&E) slides~\cite{shi2023generalist}.

Furthermore, the availability of millions of annotated H\&E images has enabled the field of computational pathology (CPath) to develop and fine-tune numerous general foundation models for diagnosis~\cite{xu2024whole,kohane2023digital} and prognostic prediction~\cite{huang2023artificial}. However, because OCT and RCM images are visually different from H\&E images~\cite{eva2020cumulative}, these models underperform due to the domain shift problem~\cite{zhou2022domain}. This discrepancy limits the utility of CPath models for optical imaging analysis.

We propose FoundationShift, a method to apply AI foundation models from computational pathology to noninvasive image analysis without retraining. Our approach relies on the counter-intuitive observation that converting optical images (i.e. OCT and RCM) to H\&E-like images before utilizing off-the-shelf computational pathology models, significantly improves model accuracy. Domain transfer models offer a distinct advantage, as they can be trained with relatively small amounts of unlabeled data. However, domain transfer models may require optical modifications to the imaging setup during training.

Several groups have demonstrated cross modality domain transfer by converting OCT~\cite{winetraub2021oct2hist}, RCM~\cite{li2021biopsy}, photoacoustic microscopy~\cite{cao2023label}, stimulated Raman~\cite{liu2024virtual}, and two photon microscopy~\cite{park2024open} images to virtually stained H\&E images by modifying the optical setup or data acquisition methods to enable "virtual staining" (See supplementary section "Imaging Setup Modifications"). These simple adjustments enable a reliable domain transfer from optical imaging to H\&E and are the key to enable FoundationShift.

Domain transfer has been used before in areas where datasets may be too small or costly to obtain. Domain transfer has been shown to improve narrow deep-learning tasks such as de-noising~\cite{warner2024multimodal} and has mostly been demonstrated in cases where source and target domains are the same modality such as two similar instruments used in different clinics~\cite{nouri2023addressing,wang2023domain}. Cross modality domain adaptation between MRI and CT has been shown to improve some specific tasks~\cite{guan2021domain}.

In comparison to the case of MRI and CT, however, histopathology has many orders of magnitude more freely available annotated data than noninvasive optical imaging. 
The lack of freely available datasets means that generic foundation models are often under trained with optical images and thus perform poorly on those images.
Therefore, we hypothesize that using domain transfer from optical imaging to H\&E followed by a generic foundation model will yield a significant accuracy boost. We demonstrate our hypothesis using OCT and RCM images in various tasks and show significant model accuracy improvements.

As illustrated in Figure \ref{sup:fig:domain_kl_divs}, domain transfer models substantially decrease the Kullback-Leibler (KL) divergence between domain transfer images and H\&E images. Notably, the KL divergence reduction is more pronounced in the case of optical imaging compared to MRI and CT. This reduction leads to the enhanced accuracy reported in this study.

\begin{figure}[t!]
\centering
\includegraphics[width=0.8\linewidth]{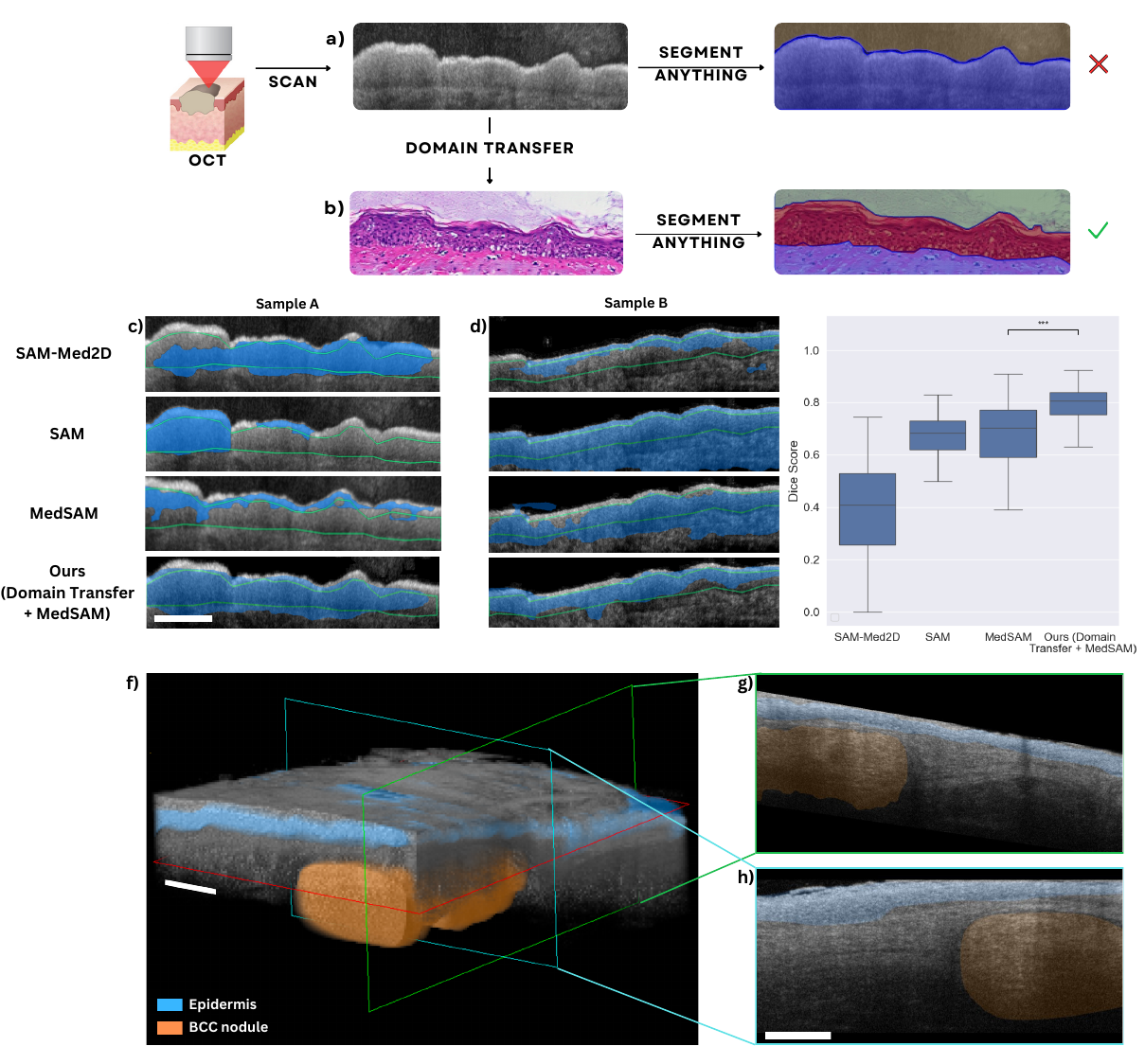}
\caption{\textbf{Tissue segmentation of OCT images utilizing FoundationShift and off-the-shelf segmentation models.}
\textbf{a)} Off-the-shelf segmentation foundation models perform poorly when segmenting OCT images. 
\textbf{b)} By performing domain transfer first, we are able to significantly improve segmentation accuracy. 
\textbf{c,d)} Segmentation examples of off-the-shelf algorithms compared to our approach (Domain Transfer + MedSAM). Ground truth expert epidermis segmentation is highlighted in green contour. Algorithm segmentation is visualized in blue.
\textbf{e)} Dice similarity coefficient quantification for each algorithm. The center line within the colored box represents the median value, with the bottom and top bounds of the box delineating the 25th and 75th percentiles, respectively, whiskers represent minimum and maximum scores over the \numberofsectionsinstudy sections. Domain transfer followed by MedSAM outperforms all off-the-shelf algorithms tested ($~p<2\cdot10^{-15}$).
\textbf{f)} Domain transfer + MedSAM  perform 3D segmentation of skin sample containing a Basal Cell Carcinoma nodule (orange). Epidermis segmentation visualized in blue.
\textbf{g,h)} 2D cross-sectional views from within the volume, which can be chosen to slice across any plane.
Scale bars in \textbf{c,f,h)} are \SI{200}{\micro\metre}.
}
\label{fig:fig1_oct_vs_vhist_seg}
\end{figure}

We first demonstrate FoundationShift utilizing OCT images. We collected a \numberofsubjectsinstudy skin samples yielding a total of \numberofsectionsinstudy paired OCT and H\&E skin images from the arm and face regions. We utilized OCT2Hist~\cite{winetraub2021oct2hist}, to convert OCT images to H\&E-like images prior to off-the-shelf segmentation models. A few examples from the dataset are shown in Figure \ref{fig:vhist_examples}.

We compared the accuracy of three off-the-shelf segmentation foundation models. Segment Anything (SAM~\cite{kirillov2023segment}) is a generic segmentation model that was trained on 11M generic images and is considered a task-agnostic model. 
In addition, MedSAM~\cite{ma2024segment} was tuned by Ma \etal utilizing 1.5M medical images, with 9.3K pathology images and 800 OCT images. Finally, SAM-Med2D~\cite{cheng2023sam} was tuned utilizing 330K pathology images and no OCT images. We hypothesized that FoundationShift would improve segmentation model accuracy, and that the improvement would correlate with number of pathology images in segmentation model training set.
As expected, all models experience Dice score increase when utilizing domain transfer (Figure \ref{sup:fig:foundation_shift_helps_all_models}), and Domain transfer + SAM-Med2D significantly outperforms all tested algorithms (Figure \ref{fig:fig1_oct_vs_vhist_seg}e).

Notably, as OCT is a 3D imaging modality, FoundationShift offers a way to extend CPath from 2D sections to a 3D volume. In Figure \ref{fig:fig1_oct_vs_vhist_seg}f, we demonstrate creating a 3D contour of segmented epidermis, and a Basal Cell Carcinoma (BCC) nodule.

\begin{figure}[t!]
\centering
\includegraphics[width=0.8\linewidth]{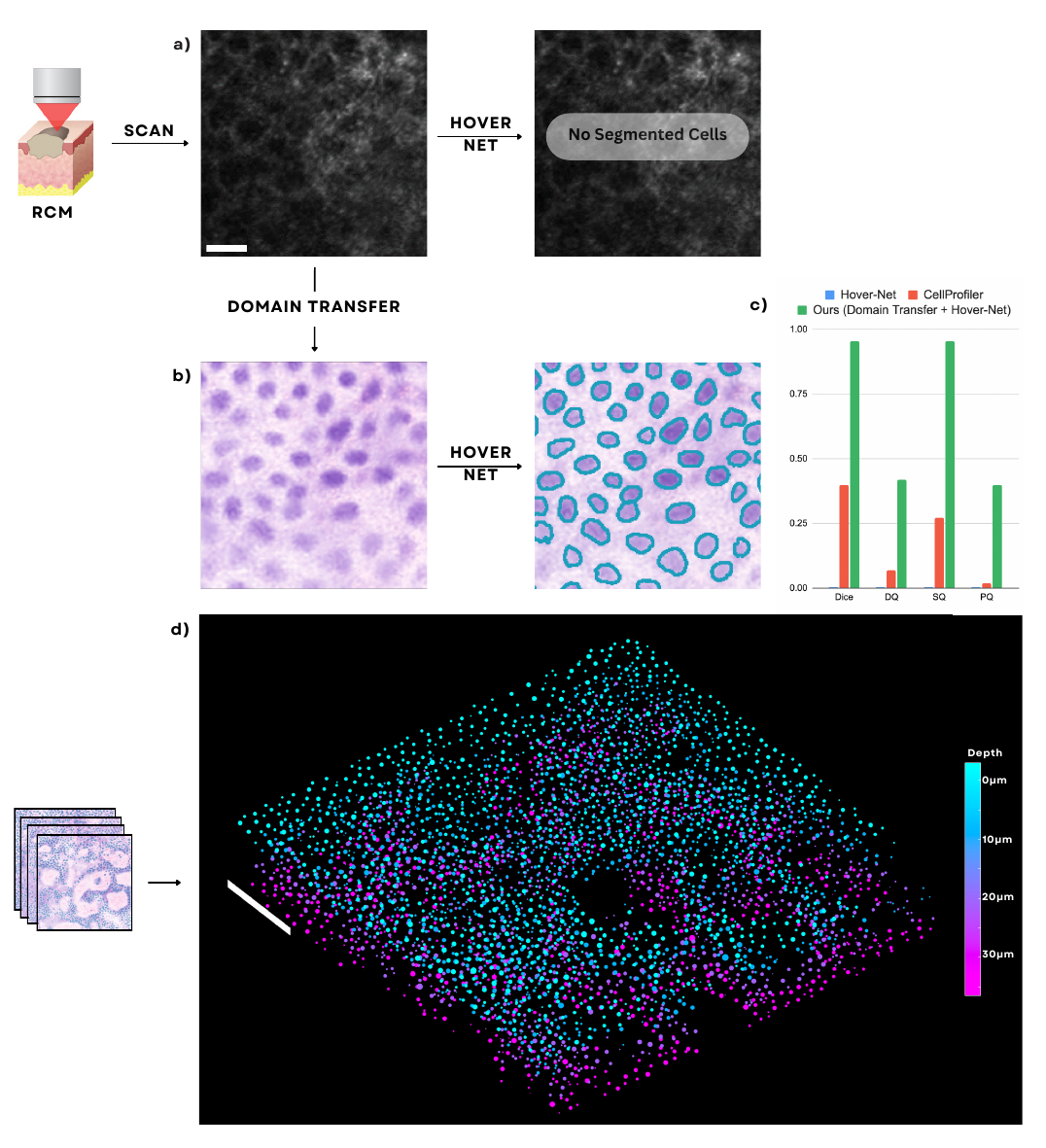}
\caption{\textbf{
Cell segmentation of RCM images utilizing FoundationShift and off-the-shelf  model.}
\textbf{a)} An RCM image obtained by Li \etal Cells are very difficult to visualize. Off-the-shelf cell segmentation Hover-Net is unable to segment any cells. 
\textbf{b)} By performing domain transfer first, we are able to significantly improve segmentation accuracy. We are able to reliably segment cells without any model tuning (zero shot).
\textbf{c)} Comparing cell segmentation accuracy of three models: Hover-Net evaluated on RCM images, CellProfiler, and our method. Our method outperforms other methods in all quality parameters: Dice score, detection quality (DQ), segmentation quality (SQ), and panoptic quality (PQ).
\textbf{d)} Domain transfer + Hover-Net perform 3D segmentation of cells in Epidermis.
Scale bar in \textbf{a)} is \SI{10}{\micro\metre}. Scale bar in \textbf{d)} is \SI{50}{\micro\metre}.
}
\label{fig:fig2_rcm_vs_vhist_seg}
\end{figure}

FoundationShift can be generalized to other noninvasive optical imaging modalities such as RCM and for fine-tuned tasks such as cell segmentation (rather then tissue segmentation demonstrated above). We utilize FoundationShift to create cell segmentation model for RCM images. To the best of our knowledge, no cell segmentation model for RCM images exists that operates without the need for any tunable parameters (zero-shot).
Li \etal~\cite{li2021biopsy} developed a domain transfer model to convert RCM images to H\&E-like images. We combined Li \etal's model with Hover-Net~\cite{graham2019hover}, a convolutional neural network designed for cell segmentation. Hover-Net was trained on 24,000 annotated cells from H\&E sections. 
For this study, we obtained four high resolution RCM images from Li \etal and employed Hover-Net to segment the cells (Figure \ref{fig:fig2_rcm_vs_vhist_seg}a and b). 
We compared cell segmentation models (Hover-Net, CellProfiler~\cite{carpenter2006cellprofiler}, and our method) against ground truth annotations over \numberofcellsinrcm cells (Figure \ref{fig:fig2_rcm_vs_vhist_seg}c). As anticipated, combining domain transfer with the off-the-shelf Hover-Net outperformed other models. This result is particularly surprising since CellProfiler has tunable parameters that can enhance accuracy, whereas Hover-Net was evaluated without any tunable parameters (zero shot). Additionally, evaluating Hover-Net on RCM images without domain transfer did not produce any cell segmentation. This outcome is expected since Hover-Net was initially trained on H\&E images.

We further evaluated FoundationShift approach on other common visual–language and large language foundation models.

In recent work, Huang \etal introduced PLIP~\cite{huang2023visual}, a visual-language model which showed notable performances in image classification of pathology H\&E slides. 
To the best of our knowledge, no visual–language foundation model exist for OCT images.
Here we introduce the first visual–language foundation model for OCT images by combining domain transfer applied with PLIP. 
We show an increase in image recall@5, which measures the number of images retrieved from the correct class out of five retrieved images.
In Figure \ref{sup:fig:oct_skin_plip} we report an increase of image recall@5 from 19\% (PLIP) to 93\% (domain transfer + PLIP) over \numberofsectionsinstudy images. Recall@5 of domain transfer + PLIP is comparable to ground truth H\&E images (82\%).

Generative Pre-Trained Transformers such as GPT-4o~\cite{achiam2023gpt} have greatly improved our ability to analyze visual information and provide automated insights. In Figure \ref{sup:fig:oct_skin_chatgpt} we report a significant improvement in ChatGPT's ability to correctly describe an OCT image. ChatGPT fails to correctly identify an OCT image of skin in 100\% of the examples. In 50\% of tests, ChatGPT incorrectly describes the skin image as a retina image. When applying FoundationShift (domain transfer + GPT-o4), the model is able to identify tissue correctly in more than 80\% of the tested images.

Finally, we evaluated FoundationShift on an image classification task often performed in off-the-shelf computational pathology models. UNI\cite{chen2024towards} was pre-trained on more than 100 million histology images to generate general-purpose self-supervised model. To be best of our knowledge, no such dataset exists for RCM. We utilized FoundationShift (domain transfer + UNI) to classify BCC nodules in RCM image from a query patch. In Figure \ref{sup:fig:rcm_skin_uni} we report qualitative improvement in the ability to identify BCC nodules in an RCM image.

Our FoundationShift approach can extend CPath AI models to noninvasive OCT and RCM without using any annotated datasets. We show a significant reduction of required data compared to model tuning approaches which requires hundreds of thousands to millions of images.
FoundationShift significantly enhances the accuracy of both generic models (SAM, ChatGPT) as well as domain specific models (MedSAM and SAM-Med2D, Hover-Net, PLIP, UNI).
Furthermore, we were able create models for unsolved image analysis tasks in OCT and RCM by combining domain transfer with off-the-shelf CPath without using any annotated dataset.

It is highly likely that FoundationShift is applicable to many noninvasive optical imaging tools and tissue types. Recently, noninvasive virtual staining models have been developed for brain (imaged by photoacoustic microscopy~\cite{kang2022deep}, stimulated Raman~\cite{liu2024virtual}), lung (imaged by autofluorescence microscopy~\cite{shithickv}) and bone (imaged by ultraviolet photoacoustic microscopy~\cite{cao2023label}). Our approach could utilize these virtual staining models to further extend computational pathology to multiple imaging modalities without the need to fine tune models for each imaging modality and body region.

Since OCT and RCM are 3D imaging modalities, FoundationShift could leverage 3D insights to improve clinical utility. It has been shown that 3D computational pathology AI enhances clinical  accuracy~\cite{song2024analysis}. In this work we demonstrated a proof of concept 3D segmentation utilizing domain transfer + CPath. As a next step, we could utilize FoundationShift to analyze 3D OCT and RCM datasets using 3D pathology models for improved clinical accuracy.

We note a few limitations of FoundationShift. Our method is inherently limited by both the accuracy of the domain transfer model as well as the CPath foundation models.

As noted in Winetraub \etal~\cite{winetraub2021oct2hist} and others, domain transfer in not an absolute representation of the underlying histology and is limited by the optical resolution of the imaging device. As a result, we do not expect FoundationShift to perform well on CPath models that require higher magnification or resolution. As better resolution high fidelity domain transfer models become available, we expect FoundationShift become more effective. In addition, CPath models discussed in this work are known to underperform human benchmark when evaluated on ground truth H\&E images. As CPath field progresses, we expect accuracy of models to increase.

Furthermore, foundation models are less accurate  compared to pre-trained domain specific AI models. For example, we expect state of the art pre trained OCT model for segmenting epidermis~\cite{del2020automatic} to outperform our generic model.
However, as task complexity increases, OCT data availability and annotation limitations will be prohibitively challenging to train domain specific models for each modality. Thus the FoundationShift approach could speed up adaption of new foundation models from CPath and across multiple domains.

Another limitation is that training a domain transfer network requires some level of effort. However, it utilizes only a small amount of non-annotated data. Domain transfer networks are typically trained utilizing up to a few thousand images~\cite{winetraub2021oct2hist, cheng2024enhanced, li2021biopsy}. 
This is significantly smaller than the typical dataset used to fine tune a foundation model. Furthermore, a single domain transfer model could boost performances of a variety of foundation models in the train optical imaging domain. Thus offering an efficient way to apply AI models from computational pathology to noninvasive optical imaging.

FoundationShift could be used to speed up noninvasive imaging research and extend computational pathology beyond H\&E. This has the potential to support clinical decision-making by identifying tumor margins or confirming tumor diagnosis.

\section*{Methods}

\subsection*{OCT and H\&E Data Acquisition}

\numberofsubjectsinstudy subjects participated in this study under IRB-24307 (ages 45 to 93, mean age of 73, 47\% female). These patients underwent Mohs surgery at Stanford's dermatology department and donated tissue discards for this study. 

Sample collection, scan and processing was descried in details previously~\cite{winetraub2021oct2hist}. We provide a short summary below.

All samples, received daily and stored in a keratinocyte media to maintain cell viability, were processed within 4 hours of excision. Each sample was trimmed to approximately 0.5cm before being encased in a fluorescent gel. This gel, consisting of gelatin and Alexa 680-NHS Ester dye, was prepared and solidified around the tissue to facilitate imaging. The OCT imaging was performed using a Thorlabs Ganymede OCT (Thorlabs GAN220) system equipped with a 10X long-working-distance objective lens (Olympus UMPLFLN10XW). This setup, with an effective numerical aperture of 0.2 and silicone oil immersion, provided a theoretical lateral resolution of 3.7 microns and an axial resolution of 2 microns.
The imaging process involved capturing 48 volume scans of 1 x 1 mm, with a 10 µm step between scans to compile a comprehensive OCT volume with isotropic lateral resolution. Post-imaging, a 650 nm laser diode was used to photobleach a barcode pattern on the tissue, which was later visualized in an overview scan of 7x8 mm to aid in tissue sectioning. Following imaging, the gel-encased samples were fixed in formalin and processed through a histological protocol to yield 15 unstained sections per tissue. These sections were subsequently stained with H\&E and scanned at 20X magnification to capture the barcode pattern and histological features.
OCT and H\&E images were aligned using an in-house registration algorithm. The OCT images, cropped to 1024 x 512 pixels at 10X magnification (1 µm/pixel), were compared with the H\&E sections. Areas with inadequate OCT signal were masked in accordance to OCT2Hist requirements~\cite{winetraub2021oct2hist}, we made sure to only use OCT sections where dermal epidermal junction (DEJ) is visible in the OCT image yielding a total of \numberofsectionsinstudy OCT images.

\subsection*{OCT Ground Truth Segmentation}

Epidermis segmentation was done by an OCT technician expert in accordance with literature guidelines~\cite{park2014skin,taghavikhalilbad2017semi}. Epidermis was segmented to exclude the stratum corneum.

Whenever there was uncertainty regarding the position of the DEJ in the OCT image, the expert was able to view the corresponding H\&E image that was precisely co-registered with the OCT image to a precision of 25 microns (see Winetraub \etal~\cite{winetraub2021oct2hist} for more details) and make a refined judgment based on the H\&E image.

Two independent OCT experts segmented the epidermis from OCT images. We then computed the Dice score between the two segmented sets, and concluded a Dice score of 0.85. This score indicates that OCT image quality is sufficient and allows for reliable segmentation of epidermis.

In skin areas near hair follicles, the epidermis often appears to stretch and envelop the follicle. Consequently, the dermal-epidermal junction in these regions is more challenging to segment. To address this issue, experts marked areas where the epidermis curves around a hair follicle with a "don't care" label. These marked regions were then excluded from the analysis.
As with epidermis, experts could consult the corresponding H\&E image.

\subsection*{OCT Tissue Segmentation and Statistical Analysis}
The segmentation pipeline consists of two main components: a model that performs domain transfer from OCT to H\&E-like images (OCT2Hist), and a segmentation model generic (SAM) or domain specific (MedSAM, SAM-Med2D). All models were applied in a zero-shot manner "as is", without any fine tuning. 
The pipeline accepts an OCT image and a bounding box prompt and outputs epidermis annotation described above. The predicted set consists of pixels identified as epidermis. We utilized bounding box prompt as it is supported by all segmentation models.

The bounding box used for segmentation was derived from expert annotations, with a height factor adjustment of 1.15, and served as input to the segmentation model. The multi-mask feature was disabled, using only the highest-confidence mask for segmentation.

Postprocessing followed the standard SAM protocol: filling small holes in the predicted mask and applying a boolean closing operation. The final mask was used directly for downstream predictions, with no further postprocessing required to project results back to the original OCT domain.

To generate a 3D segmented volume, each individual OCT slice was segmented independently and subsequently stacked to reconstruct the full volume. A 3D Gaussian with 5 pixels filter was applied to the segmentation mask.

We evaluated the segmentation accuracy of three models. Neither the segmentation models nor the domain transfer model had been exposed to any of the test data prior to evaluation. For each sample, we assessed the quality of epidermis segmentation using the Dice-Sørensen coefficient (DSC), commonly referred to as the Dice score.

The Dice score is calculated as:

\[
\text{DSC}(P, A) = \frac{2 \times |P \cap A|}{|P| + |A|}
\]

where \(P\) denotes the set of pixels in the predicted segmentation, and \(A\) represents the set of pixels in the ground truth annotation. The term \(|P \cap A|\) indicates the number of pixels (1 micron per pixel) common to both the predicted and ground truth segmentations. The Dice score ranges from 0 to 1, with a value of 1 signifying perfect overlap between the predicted and ground truth segmentation.
We further assessed the statistical significance of domain transfer improvements using the Wilcoxon signed-rank test. For each section, we measured Dice score for segmenting OCT and segmenting the virtual histology and test whether there is a significant difference between their distributions. The effect of domain transfer was verified across all three models, as shown in Figure \ref{sup:fig:foundation_shift_helps_all_models}. 

\subsection*{RCM Cell Segmentation and Statistical Analysis}
We obtained four RCM skin images and corresponding virtual H\&E images from Li \etal~\cite{li2021biopsy}. We reprocessed the images to segment the cells. Initially, we evaluated Hover-Net~\cite{graham2019hover} using PanNuke-trained weights "as is" on the RCM images. Subsequently, we used CellProfiler~\cite{carpenter2006cellprofiler} on grayscale-inverted RCM images, needing to fine-tune the CellProfiler parameters to correctly capture most visible cells. Finally, we evaluated Hover-Net on the virtual H\&E images from Li \etal.

To evaluate the accuracy of the algorithms, we established a ground truth by implementing the method from Kumar \etal~\cite{7872382}. First, we ran Hover-Net on the virtually stained dataset, then manually adjusted the results to create a ground truth consisting of \numberofcellsinrcm cells. We acknowledge that generating ground truth in this method may skew results in favor of Hover-Net. To measure potential ground truth bias, we repeated the experiment on a smaller section containing 50 cells and used CellProfiler to create a ground truth, which we then adjusted manually. We found the Dice score between the two ground truths to be 0.9. We therefore conclude that our method of annotating ground truth doesn't introduce a large bias favoring Hover-Net.

In order to evaluate cell segmentation accuracy, we utilized detection quality (DQ), segmentation quality (SQ) and panoptic quality (PQ). These were computed using Graham \etal~\cite{graham2019hover}.

\subsection*{Software Used}
We employed Roboflow version 1.1.14 for ground truth annotation. Segmentation and image processing were performed on an 8-core CPU, while domain transfer was handled by a 10-core GPU, both supported by the Apple M2 chip. The entire pipeline was written in Python (3.10.13) using the following external packages: numpy (1.26.2), opencv-python (4.8.1.78), pillow (10.1.0), pandas (2.1.4), scikit-learn (1.3.2), pip (23.3.1), seaborn (0.13.2), scikit-image (0.22.0), and torchvision (0.16.2). We used Pytorch (2.1.2) for deep learning. All plots were generated using matplotlib (3.8.2) and seaborn (0.13.2), and edited using Canva and Adobe Photoshop. ImageJ (1.54f) with the volume viewer plugin visualized the 3D volume in figure 4. Segmentation was done using the March '24 git versions shared by SAM, MedSAM, and SAM-Med2D authors in their respective GitHub repositories.

\section*{Acknowledgements}
We would like to thank Paulo de Luna, Patricio Gonzalez Roa, Joshua Vera, Kent Lee and Jahanbanoo Shahryari for their comments, support and contributions.

\textbf{Funding}: This work has been supported by National Institutes of Health grant NIH DP5OD031858; Stanford Office of Technology Licensing grant 323728; Stanford Bio-X interdisciplinary initiative seed grant IIP11-9; EB Research Partnership grant SPO 315482; Sanofi S.A. research grant SPO 339163; the Stanford Bowes Bio-X Graduate Fellowship Stanford Biophysics Program training
grant T32GM-08294; 

\textbf{Author contributions}:
Y.W. conceived of the presented idea, developed the theory behind the model shift. D.B., O.H., W.K.A., performed the computation of the domain transfer. D.B. developed the statistical tests.
E.M., A.V., S.A., collected samples
from patients, imaged and processed samples to generate the dataset used in this paper.
J.L., P.O.S. and A.O. have collected RCM dataset and created a virtual staining of RCM images.
D.B., E.M. annotated the dataset.
Y.W., O.F. and K.S., K.R. contributed to the overall design and direction of the research. the manuscript was written through contributions of all authors. Authors have
given approval to the final version of the manuscript.

\textbf{Competing interests}: Authors declare no competing interests.

\textbf{Data availability}: All data needed to evaluate the conclusions in the paper are present in the paper and/or the Supplementary Materials. 

\bibliography{main}

\begin{thebibliography}{10}
\urlstyle{rm}
\expandafter\ifx\csname url\endcsname\relax
  \def\url#1{\texttt{#1}}\fi
\expandafter\ifx\csname urlprefix\endcsname\relax\def\urlprefix{URL }\fi
\expandafter\ifx\csname doiprefix\endcsname\relax\def\doiprefix{DOI: }\fi
\providecommand{\bibinfo}[2]{#2}
\providecommand{\eprint}[2][]{\url{#2}}

\bibitem{fisher2018clinical}
\bibinfo{author}{Fisher, J.}, \bibinfo{author}{Siegel, D.~M.} \& \bibinfo{author}{Markowitz, O.}
\newblock \bibinfo{journal}{\bibinfo{title}{Clinical utility of bedside multibeam optical coherence tomography imaging in a patient with multiple basal cell carcinomas}}.
\newblock {\emph{\JournalTitle{Dermatologic Surgery}}} \textbf{\bibinfo{volume}{44}}, \bibinfo{pages}{874--876} (\bibinfo{year}{2018}).

\bibitem{ulrich2015sensitivity}
\bibinfo{author}{Ulrich, M.} \emph{et~al.}
\newblock \bibinfo{journal}{\bibinfo{title}{The sensitivity and specificity of optical coherence tomography for the assisted diagnosis of nonpigmented basal cell carcinoma: an observational study}}.
\newblock {\emph{\JournalTitle{British Journal of Dermatology}}} \textbf{\bibinfo{volume}{173}}, \bibinfo{pages}{428--435} (\bibinfo{year}{2015}).

\bibitem{li2014high}
\bibinfo{author}{Li, G.} \emph{et~al.}
\newblock \bibinfo{journal}{\bibinfo{title}{High-definition optical coherence tomography in the diagnosis of basal cell carcinoma evaluated by an experienced versus inexperienced investigator}}.
\newblock {\emph{\JournalTitle{Journal of Clinical \& Experimental Dermatology Research}}} \textbf{\bibinfo{volume}{5}}, \bibinfo{pages}{1--4} (\bibinfo{year}{2014}).

\bibitem{kuck2014evaluation}
\bibinfo{author}{Kuck, M.} \emph{et~al.}
\newblock \bibinfo{journal}{\bibinfo{title}{Evaluation of optical coherence tomography as a non-invasive diagnostic tool in cutaneous wound healing}}.
\newblock {\emph{\JournalTitle{Skin Research and Technology}}} \textbf{\bibinfo{volume}{20}}, \bibinfo{pages}{1--7} (\bibinfo{year}{2014}).

\bibitem{tsai2021submicron}
\bibinfo{author}{Tsai, C.-Y.} \emph{et~al.}
\newblock \bibinfo{journal}{\bibinfo{title}{Submicron spatial resolution optical coherence tomography for visualising the 3d structures of cells cultivated in complex culture systems}}.
\newblock {\emph{\JournalTitle{Scientific Reports}}} \textbf{\bibinfo{volume}{11}}, \bibinfo{pages}{3492} (\bibinfo{year}{2021}).

\bibitem{lin2024rapid}
\bibinfo{author}{Lin, C.-H.} \emph{et~al.}
\newblock \bibinfo{journal}{\bibinfo{title}{Rapid measurement of epidermal thickness in oct images of skin}}.
\newblock {\emph{\JournalTitle{Scientific Reports}}} \textbf{\bibinfo{volume}{14}}, \bibinfo{pages}{2230} (\bibinfo{year}{2024}).

\bibitem{kumar2023deep}
\bibinfo{author}{Kumar, P.}, \bibinfo{author}{Dhara, S.}, \bibinfo{author}{Gope, A.}, \bibinfo{author}{Chatterjee, J.} \& \bibinfo{author}{Mandal, S.}
\newblock \bibinfo{title}{Deep learning based skin-layer segmentation for characterizing cutaneous wounds from optical coherence tomography images}.
\newblock In \emph{\bibinfo{booktitle}{2023 45th Annual International Conference of the IEEE Engineering in Medicine \& Biology Society (EMBC)}}, \bibinfo{pages}{1--4} (\bibinfo{organization}{IEEE}, \bibinfo{year}{2023}).

\bibitem{shi2023generalist}
\bibinfo{author}{Shi, P.} \emph{et~al.}
\newblock \bibinfo{journal}{\bibinfo{title}{Generalist vision foundation models for medical imaging: A case study of segment anything model on zero-shot medical segmentation}}.
\newblock {\emph{\JournalTitle{Diagnostics}}} \textbf{\bibinfo{volume}{13}}, \bibinfo{pages}{1947} (\bibinfo{year}{2023}).

\bibitem{xu2024whole}
\bibinfo{author}{Xu, H.} \emph{et~al.}
\newblock \bibinfo{journal}{\bibinfo{title}{A whole-slide foundation model for digital pathology from real-world data}}.
\newblock {\emph{\JournalTitle{Nature}}} \bibinfo{pages}{1--8} (\bibinfo{year}{2024}).

\bibitem{kohane2023digital}
\bibinfo{author}{Kohane, I.~S.}, \bibinfo{author}{Churchill, S.}, \bibinfo{author}{Tan, A. L.~M.}, \bibinfo{author}{Vella, M.} \& \bibinfo{author}{Perry, C.~L.}
\newblock \bibinfo{journal}{\bibinfo{title}{The digital--physical divide for pathology research}}.
\newblock {\emph{\JournalTitle{The Lancet Digital Health}}} \textbf{\bibinfo{volume}{5}}, \bibinfo{pages}{e859--e861} (\bibinfo{year}{2023}).

\bibitem{huang2023artificial}
\bibinfo{author}{Huang, Z.} \emph{et~al.}
\newblock \bibinfo{journal}{\bibinfo{title}{Artificial intelligence reveals features associated with breast cancer neoadjuvant chemotherapy responses from multi-stain histopathologic images}}.
\newblock {\emph{\JournalTitle{NPJ Precision Oncology}}} \textbf{\bibinfo{volume}{7}}, \bibinfo{pages}{14} (\bibinfo{year}{2023}).

\bibitem{eva2020cumulative}
\bibinfo{author}{Eva, V.} \emph{et~al.}
\newblock \bibinfo{journal}{\bibinfo{title}{Cumulative sum analysis for the learning curve of optical coherence tomography assisted diagnosis of basal cell carcinoma}}.
\newblock {\emph{\JournalTitle{Acta dermato-venereologica}}} \textbf{\bibinfo{volume}{100}} (\bibinfo{year}{2020}).

\bibitem{zhou2022domain}
\bibinfo{author}{Zhou, K.}, \bibinfo{author}{Liu, Z.}, \bibinfo{author}{Qiao, Y.}, \bibinfo{author}{Xiang, T.} \& \bibinfo{author}{Loy, C.~C.}
\newblock \bibinfo{journal}{\bibinfo{title}{Domain generalization: A survey}}.
\newblock {\emph{\JournalTitle{IEEE Transactions on Pattern Analysis and Machine Intelligence}}} \textbf{\bibinfo{volume}{45}}, \bibinfo{pages}{4396--4415} (\bibinfo{year}{2022}).

\bibitem{winetraub2021oct2hist}
\bibinfo{author}{Winetraub, Y.} \emph{et~al.}
\newblock \bibinfo{journal}{\bibinfo{title}{Noninvasive virtual biopsy using micro-registered optical coherence tomography (oct) in human subjects}}.
\newblock {\emph{\JournalTitle{Science Advances}}} \textbf{\bibinfo{volume}{10}}, \bibinfo{pages}{eadi5794} (\bibinfo{year}{2024}).

\bibitem{li2021biopsy}
\bibinfo{author}{Li, J.} \emph{et~al.}
\newblock \bibinfo{journal}{\bibinfo{title}{Biopsy-free in vivo virtual histology of skin using deep learning}}.
\newblock {\emph{\JournalTitle{Light: Science {\&} Applications}}} \textbf{\bibinfo{volume}{10}}, \bibinfo{pages}{233} (\bibinfo{year}{2021}).

\bibitem{cao2023label}
\bibinfo{author}{Cao, R.} \emph{et~al.}
\newblock \bibinfo{journal}{\bibinfo{title}{Label-free intraoperative histology of bone tissue via deep-learning-assisted ultraviolet photoacoustic microscopy}}.
\newblock {\emph{\JournalTitle{Nature biomedical engineering}}} \textbf{\bibinfo{volume}{7}}, \bibinfo{pages}{124--134} (\bibinfo{year}{2023}).

\bibitem{liu2024virtual}
\bibinfo{author}{Liu, Z.} \emph{et~al.}
\newblock \bibinfo{journal}{\bibinfo{title}{Virtual formalin-fixed and paraffin-embedded staining of fresh brain tissue via stimulated raman cyclegan model}}.
\newblock {\emph{\JournalTitle{Science Advances}}} \textbf{\bibinfo{volume}{10}}, \bibinfo{pages}{eadn3426} (\bibinfo{year}{2024}).

\bibitem{park2024open}
\bibinfo{author}{Park, W.~Y.} \emph{et~al.}
\newblock \bibinfo{journal}{\bibinfo{title}{Open-top bessel beam two-photon light sheet microscopy for three-dimensional pathology}}.
\newblock {\emph{\JournalTitle{Elife}}} \textbf{\bibinfo{volume}{12}}, \bibinfo{pages}{RP92614} (\bibinfo{year}{2024}).

\bibitem{warner2024multimodal}
\bibinfo{author}{Warner, E.} \emph{et~al.}
\newblock \bibinfo{journal}{\bibinfo{title}{Multimodal machine learning in image-based and clinical biomedicine: Survey and prospects}}.
\newblock {\emph{\JournalTitle{International Journal of Computer Vision}}} \bibinfo{pages}{1--17} (\bibinfo{year}{2024}).

\bibitem{nouri2023addressing}
\bibinfo{author}{Nouri, H.}, \bibinfo{author}{Nasri, R.} \& \bibinfo{author}{Abtahi, S.-H.}
\newblock \bibinfo{journal}{\bibinfo{title}{Addressing inter-device variations in optical coherence tomography angiography: will image-to-image translation systems help?}}
\newblock {\emph{\JournalTitle{International Journal of Retina and Vitreous}}} \textbf{\bibinfo{volume}{9}}, \bibinfo{pages}{51} (\bibinfo{year}{2023}).

\bibitem{wang2023domain}
\bibinfo{author}{Wang, J.}, \bibinfo{author}{Zong, Y.}, \bibinfo{author}{He, Y.}, \bibinfo{author}{Shi, G.} \& \bibinfo{author}{Jiang, C.}
\newblock \bibinfo{journal}{\bibinfo{title}{Domain adaptation-based automated detection of retinal diseases from optical coherence tomography images}}.
\newblock {\emph{\JournalTitle{Current Eye Research}}} \textbf{\bibinfo{volume}{48}}, \bibinfo{pages}{836--842} (\bibinfo{year}{2023}).

\bibitem{guan2021domain}
\bibinfo{author}{Guan, H.} \& \bibinfo{author}{Liu, M.}
\newblock \bibinfo{journal}{\bibinfo{title}{Domain adaptation for medical image analysis: a survey}}.
\newblock {\emph{\JournalTitle{IEEE Transactions on Biomedical Engineering}}} \textbf{\bibinfo{volume}{69}}, \bibinfo{pages}{1173--1185} (\bibinfo{year}{2021}).

\bibitem{kirillov2023segment}
\bibinfo{author}{Kirillov, A.} \emph{et~al.}
\newblock \bibinfo{title}{Segment anything}.
\newblock In \emph{\bibinfo{booktitle}{Proceedings of the IEEE/CVF International Conference on Computer Vision}}, \bibinfo{pages}{4015--4026} (\bibinfo{year}{2023}).

\bibitem{ma2024segment}
\bibinfo{author}{Ma, J.} \emph{et~al.}
\newblock \bibinfo{journal}{\bibinfo{title}{Segment anything in medical images}}.
\newblock {\emph{\JournalTitle{Nature Communications}}} \textbf{\bibinfo{volume}{15}}, \bibinfo{pages}{654} (\bibinfo{year}{2024}).

\bibitem{cheng2023sam}
\bibinfo{author}{Cheng, J.} \emph{et~al.}
\newblock \bibinfo{journal}{\bibinfo{title}{Sam-med2d}}.
\newblock {\emph{\JournalTitle{arXiv preprint arXiv:2308.16184}}}  (\bibinfo{year}{2023}).

\bibitem{graham2019hover}
\bibinfo{author}{Graham, S.} \emph{et~al.}
\newblock \bibinfo{journal}{\bibinfo{title}{Hover-net: Simultaneous segmentation and classification of nuclei in multi-tissue histology images}}.
\newblock {\emph{\JournalTitle{Medical image analysis}}} \textbf{\bibinfo{volume}{58}}, \bibinfo{pages}{101563} (\bibinfo{year}{2019}).

\bibitem{carpenter2006cellprofiler}
\bibinfo{author}{Carpenter, A.~E.} \emph{et~al.}
\newblock \bibinfo{journal}{\bibinfo{title}{Cellprofiler: image analysis software for identifying and quantifying cell phenotypes}}.
\newblock {\emph{\JournalTitle{Genome biology}}} \textbf{\bibinfo{volume}{7}}, \bibinfo{pages}{1--11} (\bibinfo{year}{2006}).

\bibitem{huang2023visual}
\bibinfo{author}{Huang, Z.}, \bibinfo{author}{Bianchi, F.}, \bibinfo{author}{Yuksekgonul, M.}, \bibinfo{author}{Montine, T.~J.} \& \bibinfo{author}{Zou, J.}
\newblock \bibinfo{journal}{\bibinfo{title}{A visual--language foundation model for pathology image analysis using medical twitter}}.
\newblock {\emph{\JournalTitle{Nature medicine}}} \textbf{\bibinfo{volume}{29}}, \bibinfo{pages}{2307--2316} (\bibinfo{year}{2023}).

\bibitem{achiam2023gpt}
\bibinfo{author}{Achiam, J.} \emph{et~al.}
\newblock \bibinfo{journal}{\bibinfo{title}{Gpt-4 technical report}}.
\newblock {\emph{\JournalTitle{arXiv preprint arXiv:2303.08774}}}  (\bibinfo{year}{2023}).

\bibitem{chen2024towards}
\bibinfo{author}{Chen, R.~J.} \emph{et~al.}
\newblock \bibinfo{journal}{\bibinfo{title}{Towards a general-purpose foundation model for computational pathology}}.
\newblock {\emph{\JournalTitle{Nature Medicine}}} \textbf{\bibinfo{volume}{30}}, \bibinfo{pages}{850--862} (\bibinfo{year}{2024}).

\bibitem{kang2022deep}
\bibinfo{author}{Kang, L.}, \bibinfo{author}{Li, X.}, \bibinfo{author}{Zhang, Y.} \& \bibinfo{author}{Wong, T.~T.}
\newblock \bibinfo{journal}{\bibinfo{title}{Deep learning enables ultraviolet photoacoustic microscopy based histological imaging with near real-time virtual staining}}.
\newblock {\emph{\JournalTitle{Photoacoustics}}} \textbf{\bibinfo{volume}{25}}, \bibinfo{pages}{100308} (\bibinfo{year}{2022}).

\bibitem{shithickv}
\bibinfo{author}{Shi, L.} \emph{et~al.}
\newblock \bibinfo{title}{Thickv-stain: Unprocessed thick tissues virtual staining for rapid intraoperative histology}.
\newblock In \emph{\bibinfo{booktitle}{Medical Imaging with Deep Learning}} (\bibinfo{year}{2024}).

\bibitem{song2024analysis}
\bibinfo{author}{Song, A.~H.} \emph{et~al.}
\newblock \bibinfo{journal}{\bibinfo{title}{Analysis of 3d pathology samples using weakly supervised ai}}.
\newblock {\emph{\JournalTitle{Cell}}} \textbf{\bibinfo{volume}{187}}, \bibinfo{pages}{2502--2520} (\bibinfo{year}{2024}).

\bibitem{del2020automatic}
\bibinfo{author}{Del~Amor, R.} \emph{et~al.}
\newblock \bibinfo{journal}{\bibinfo{title}{Automatic segmentation of epidermis and hair follicles in optical coherence tomography images of normal skin by convolutional neural networks}}.
\newblock {\emph{\JournalTitle{Frontiers in Medicine}}} \textbf{\bibinfo{volume}{7}}, \bibinfo{pages}{220} (\bibinfo{year}{2020}).

\bibitem{cheng2024enhanced}
\bibinfo{author}{Cheng, S.} \emph{et~al.}
\newblock \bibinfo{journal}{\bibinfo{title}{Enhanced multiscale human brain imaging by semi-supervised digital staining and serial sectioning optical coherence tomography}}.
\newblock {\emph{\JournalTitle{Research Square}}}  (\bibinfo{year}{2024}).

\bibitem{park2014skin}
\bibinfo{author}{Park, E.~S.}
\newblock \bibinfo{journal}{\bibinfo{title}{Skin-layer analysis using optical coherence tomography (oct)}}.
\newblock {\emph{\JournalTitle{Medical Lasers}}} \textbf{\bibinfo{volume}{3}}, \bibinfo{pages}{1--4} (\bibinfo{year}{2014}).

\bibitem{taghavikhalilbad2017semi}
\bibinfo{author}{Taghavikhalilbad, A.} \emph{et~al.}
\newblock \bibinfo{journal}{\bibinfo{title}{Semi-automated localization of dermal epidermal junction in optical coherence tomography images of skin}}.
\newblock {\emph{\JournalTitle{Applied optics}}} \textbf{\bibinfo{volume}{56}}, \bibinfo{pages}{3116--3121} (\bibinfo{year}{2017}).

\bibitem{7872382}
\bibinfo{author}{Kumar, N.} \emph{et~al.}
\newblock \bibinfo{journal}{\bibinfo{title}{A dataset and a technique for generalized nuclear segmentation for computational pathology}}.
\newblock {\emph{\JournalTitle{IEEE Transactions on Medical Imaging}}} \textbf{\bibinfo{volume}{36}}, \bibinfo{pages}{1550--1560}, \doiprefix\url{10.1109/TMI.2017.2677499} (\bibinfo{year}{2017}).

\bibitem{jin2019deep}
\bibinfo{author}{Jin, C.-B.} \emph{et~al.}
\newblock \bibinfo{journal}{\bibinfo{title}{Deep ct to mr synthesis using paired and unpaired data}}.
\newblock {\emph{\JournalTitle{Sensors}}} \textbf{\bibinfo{volume}{19}}, \bibinfo{pages}{2361} (\bibinfo{year}{2019}).

\bibitem{he2016deep_resnet50}
\bibinfo{author}{He, K.}, \bibinfo{author}{Zhang, X.}, \bibinfo{author}{Ren, S.} \& \bibinfo{author}{Sun, J.}
\newblock \bibinfo{title}{Deep residual learning for image recognition}.
\newblock In \emph{\bibinfo{booktitle}{Proceedings of the IEEE conference on computer vision and pattern recognition}}, \bibinfo{pages}{770--778} (\bibinfo{year}{2016}).

\end{thebibliography}
\newpage

\setcounter{figure}{0}
\renewcommand{\thefigure}{S\arabic{figure}}

\setcounter{table}{0} 
\renewcommand{\thetable}{S\arabic{table}} 

\section*{Domain Transfer Model for OCT: OCT2Hist}
\begin{figure}[H]
\centering
\includegraphics[width=0.9\linewidth]{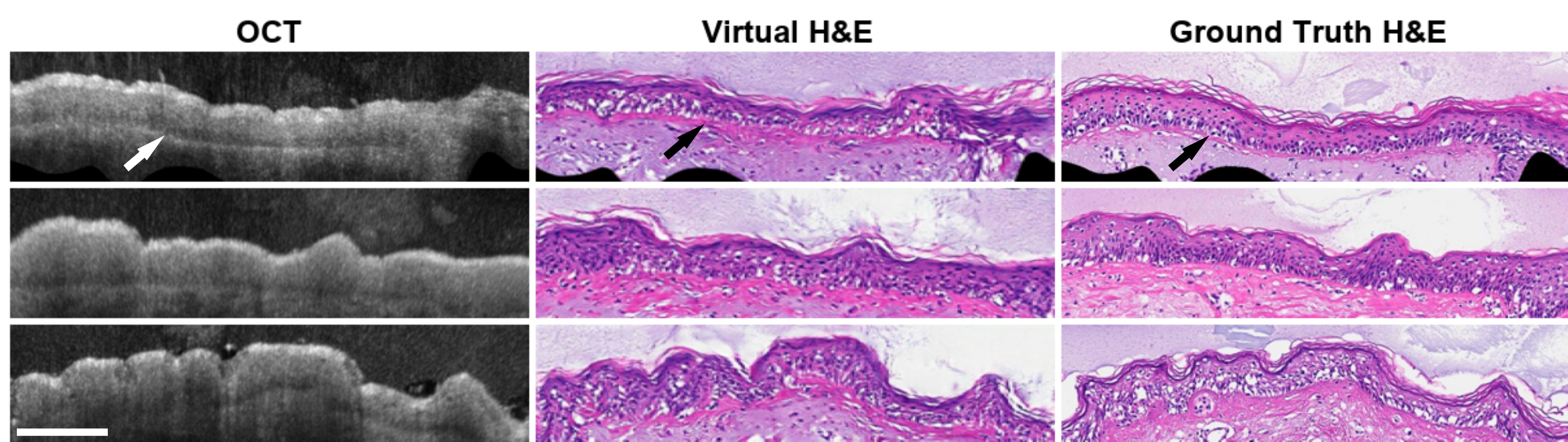}
\caption{\textbf{OCT2Hist: domain transfer from OCT to virtual H\&E.} We show examples of an OCT image (left column), the corresponding domain transfer computer generated virtual H\&E image (middle column), and the corresponding ground truth histology image (right column) from a few skin samples. Dermal epidermal junction is visible in both OCT and ground truth H\&E and is reproduced by virtual H\&E (arrows). Scale bar: \SI{200}{\micro\metre}.}
\label{fig:vhist_examples}
\end{figure}

\section*{Theoretical Discussion}
\begin{figure}[H]
\centering
\includegraphics[width=0.65\linewidth]{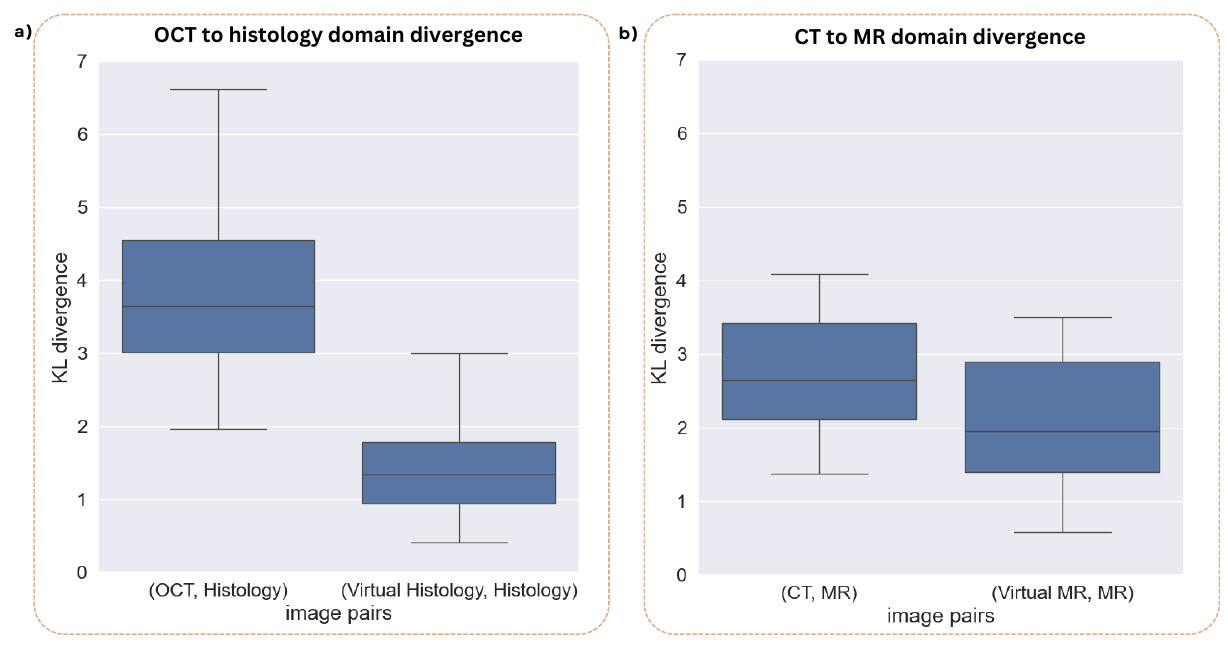}
\caption{\textbf{Measuring the Kullback Leibler (KL) divergence between OCT and histology.}
\textbf{a)} The KL divergence between OCT images and H\&E images from the same locations, before and after domain transfer. The higher KL divergence between OCT and H\&E before domain transfer indicates a greater dissimilarity, while the reduced divergence after domain transfer suggests a closer alignment to the H\&E training domain. 
\textbf{b)} A comparison of KL divergence in a related study performing domain transfer from CT to MR~\cite{jin2019deep}, showing less significant improvement compared to the OCT to H\&E domain transfer. These results highlight the substantial impact of domain transfer in aligning images closer to the training domain, enhancing the performance of computational pathology models.
}
\label{sup:fig:domain_kl_divs}
\end{figure}

We hypothesize that domain transfer (virtual H\&E) aligns the generated images much closer to the computational pathology training domain (H\&E) and improves model accuracy. 
To test this hypothesis, we computed the Kullback-Leibler (KL) divergence between OCT images and corresponding H\&E images from the same locations across \numberofsectionsinstudy images. As shown in Figure \ref{sup:fig:domain_kl_divs}a, the KL divergence between OCT and H\&E images is much higher than the divergence after domain transfer (virtual H\&E). This indicates that domain transfer significantly aligns images closer to the training domain, thus improving the performance of off-the-shelf computational pathology models.

To further validate our hypothesis, we compared the KL divergence of OCT and H\&E to a related study on domain transfer from CT to MR~\cite{jin2019deep}. As seen in Figure \ref{sup:fig:domain_kl_divs}b, domain transfer did bring the KL distance closer to ground truth MR images but with less improvement compared to OCT. This explains our dramatic results: first, images align closer to the pre-trained domain; second, the improvement is more significant compared to other domain transfers.

The KL divergence is calculated on embedding pairs of images using VGG16.

\section*{FoundationShift Boost Tissue Segmentation Accuracy for All Tested Models}

In this section we evaluate SAM, MedSAM and SAM-Med2D segmentation accuracy with and without domain transfer and compare results with ground truth H\&E segmentation.

\begin{figure}[H]
\centering
\includegraphics[width=0.45\linewidth]{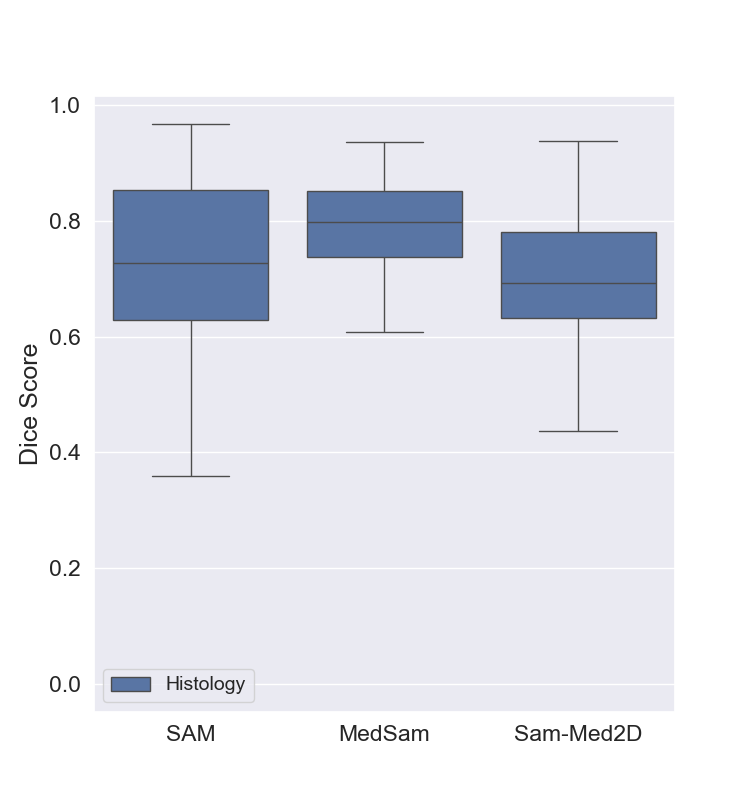}
\caption{\textbf{Segmentation model performances on ground truth H\&E data.} Dice score distribution when evaluating models on H\&E data. The center line within the colored box represents the median value, with the bottom and top bounds of the box delineating the 25th and 75th percentiles, respectively, whiskers represent minimum and maximum scores over the \numberofsectionsinstudy sections. Mean Dice scores of SAM, MedSAM, and Sam-Med2D: \textbf{0.69}, \textbf{0.78}, and \textbf{0.68}, respectively.}
\label{sup:fig:real_hist_dice}
\end{figure}

We first evaluate baseline performances of each segmentation model on H\&E images. We manually segmented epidermis in \numberofsectionsinstudy ground truth H\&E images from our dataset, and evaluated the Dice score on each model (Figure \ref{sup:fig:real_hist_dice}). The mean scores for SAM, MedSAM and SAM-Med2D were 0.69, 0.78, and 0.68 respectively.
Results are consistent with literature as SAM-Med2D is known to outperform SAM~\cite{ma2024segment}. However SAM-Med2D did not outperform SAM as expected~\cite{cheng2023sam} in this specific segmentation task.

\begin{figure}[H]
\centering
\includegraphics[width=\linewidth]{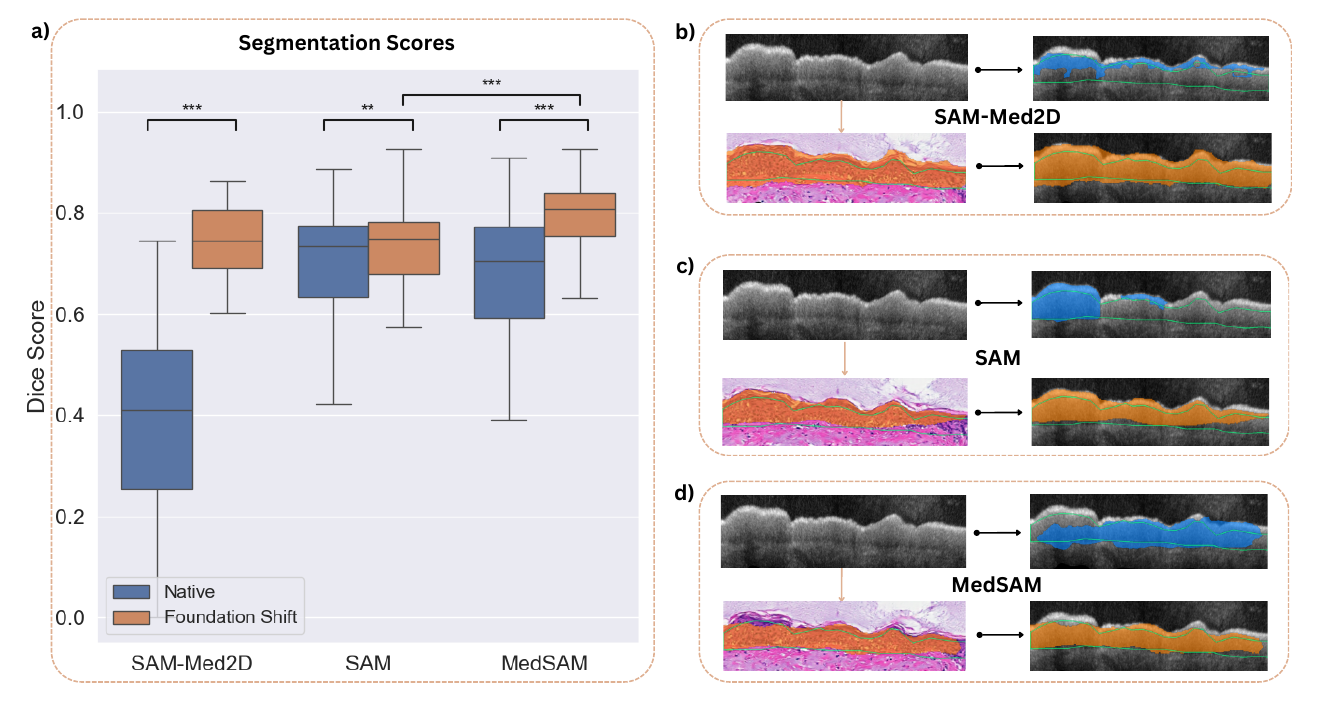}
\caption{ \textbf{Increase in Dice scores following domain transfer. a)} Dice score of the three algorithms. The center line within the colored box represents the median value, with the bottom and top bounds of the box delineating the 25th and 75th percentiles, respectively, whiskers represent minimum and maximum scores over the \numberofsectionsinstudy sections. All three algorithms (SAM-Med2D, SAM and MedSAM) showed significant Dice score increase $~p<3\cdot10^{-17},$ $p<6\cdot10^{-3},$ and $~p<2\cdot10^{-15}$  respectively (see Table \ref{sup:model_performance_statisics}). A comparison between SAM and MedSAM reveals that H\&E domain specific training has a significant effect on accuracy $p<1.57\cdot10^{-66}$. \textbf{b)-d)} An example of segmentation performances. Each panel contains the original OCT image (top left), the OCT segmented (blue, top right), segmentation after domain transfer (orange, bottom left) and the segmentation projected over the OCT image (bottom right). Comparing blue segmentation with orange segmentation in each panel, we can see that domain transfer increases accuracy. Epidermis ground truth is outlined in green.}
\label{sup:fig:foundation_shift_helps_all_models}
\end{figure}

Next, we applied domain transfer followed by SAM, MedSAM and SAM-Med2D.
Segmentation Dice scores (Figure \ref{sup:fig:foundation_shift_helps_all_models}) are consistent with ground truth H\&E accuracy mentioned above; SAM's virtual H\&E Dice score (0.69) and SAM-Med2D's Dice score (0.68) are fairly similar, however MedSAM has a significantly higher score (0.78) when compared to SAM and SAM-Med2D. 

All three algorithms showed significant Dice score increase after applying domain transfer: $p<6\cdot10^{-3},$ $~p<2\cdot10^{-15},$ and $~p<3\cdot10^{-17}$ respectively (see Table \ref{sup:model_performance_statisics}).
A comparison between SAM and MedSAM reveals that H\&E domain specific training has a significant effect on accuracy $p<1.57\cdot10^{-66}$.

On the other hand, SAM (generic model) exhibits modest yet significant score increase of 0.05 $(p<6\cdot10^{-3})$. Both MedSAM and SAM-Med2D, which were trained with pathology images, show a much larger increase in  score of 0.14 $(p<2\cdot10^{-15})$ and 0.36 $(p<3\cdot10^{-17})$, respectively.

When segmenting OCT images, both MedSAM and SAM-Med2D have moderate to low Dice scores. This observation underscores the limited generalization ability of the specialist models on unseen targets with few to zero OCT images in the training set. However, the low OCT scores are boosted above the general SAM model when performing a domain transfer. Even though SAM performed better than MedSAM on OCT images (Dice score of 0.68 and 0.67, respectively), MedSAM was significantly more accurate than SAM $(p<2\cdot10^{-9})$ after performing domain transfer (Dice score of 0.73 and 0.80, respectively), highlighting the potential performance boost specialized computational pathology algorithms could achieve with domain transfer.

Finally, we evaluated our inter-expert agreement by having two experts segment 10 representative images from the OCT dataset. Both experts followed the same annotation protocol. Inter-expert Dice score was 0.85 and serves as a bound to expert agreement.

\begin{table}[H]
    \centering
    \begin{tabular}{|c|c c|c c|c c|}
        \hline
        Model & \multicolumn{2}{c|}{SAM} & \multicolumn{2}{c|}{MedSAM} & \multicolumn{2}{c|}{SAM-Med2D} \\
        \hline
        With Domain Transfer? & No & Yes & No & Yes & No & Yes \\
        \hline
        Mean & 0.68 & 0.73 & 0.67 & 0.80 & 0.38 & 0.74 \\
        Median & 0.73 & 0.75 & 0.70 & 0.81 & 0.41 & 0.74 \\
        Std & 0.16 & 0.07 & 0.16 & 0.07 & 0.18 & 0.07 \\
        Min & 0.00 & 0.57 & 0.18 & 0.63 & 0.00 & 0.60 \\
        Max & 0.89 & 0.93 & 0.91 & 0.92 & 0.74 & 0.86 \\
        25th percentile & 0.63 & 0.68 & 0.59 & 0.75 & 0.26 & 0.69 \\
        75th percentile & 0.77 & 0.78 & 0.77 & 0.84 & 0.53 & 0.81 \\
        \hline
    \end{tabular}
    \caption{Segmentation model accuracy with (Yes) and without (No) domain transfer. Statistics over \numberofsectionsinstudy sections}
    \label{sup:model_performance_statisics}
\end{table}

\ifthenelse{\boolean{false}}{ 
\subsection*{Examples of Segmented Sections}

\def\additionalexamplescaption{
Examples of segmentation performances. Each row contains \textbf{a)} the original OCT image, \textbf{b)} the OCT segmented (blue), \textbf{c)} domain transfer followed by segmentation (orange), and \textbf{d)} the segmentation projected over the OCT image. Epidermis ground truth is outlined in green. "Don't Care" label marked as green square.
}

\begin{figure}[H]
\centering
\includegraphics[width=0.9\linewidth]{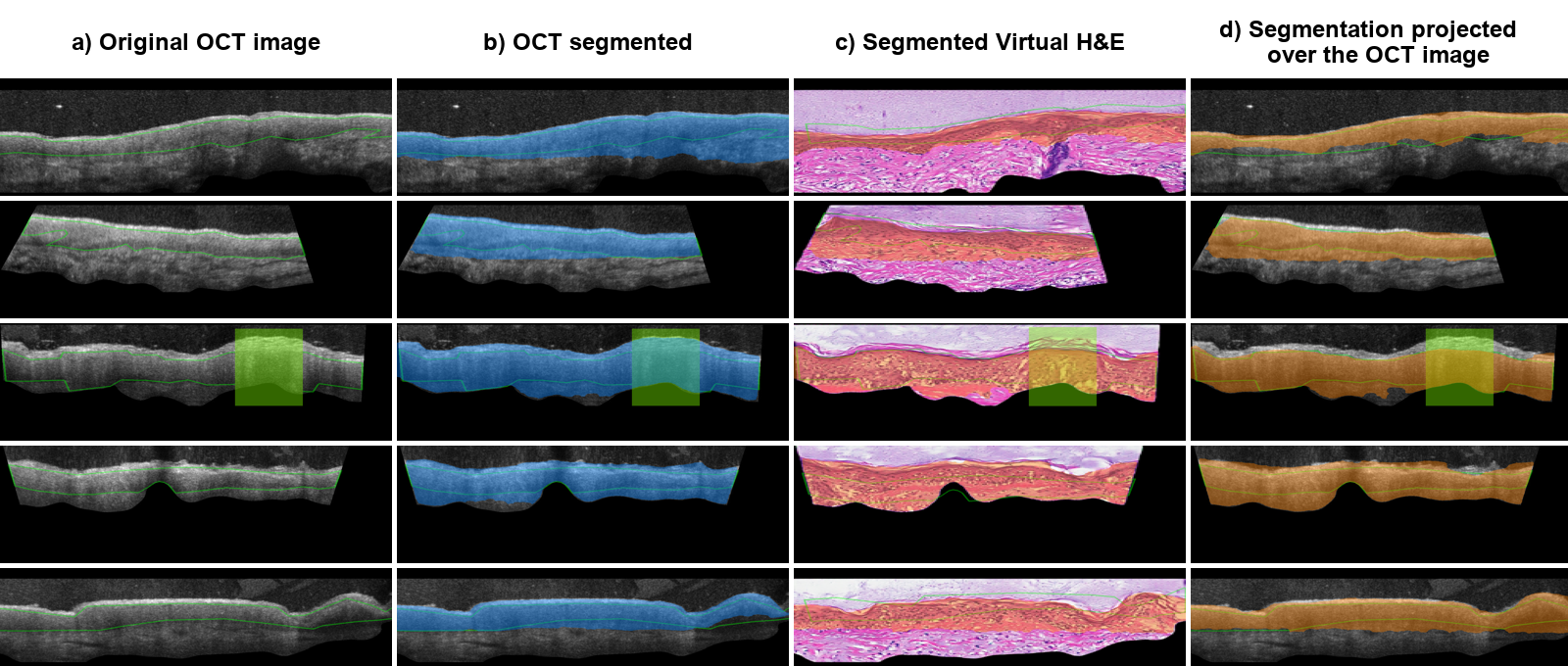}
\caption{\textbf{Segmented areas by SAM}. \additionalexamplescaption}
\label{sup:fig:more_segmented_areas_sam}
\end{figure}

\begin{figure}[H]
\centering
\includegraphics[width=0.9\linewidth]{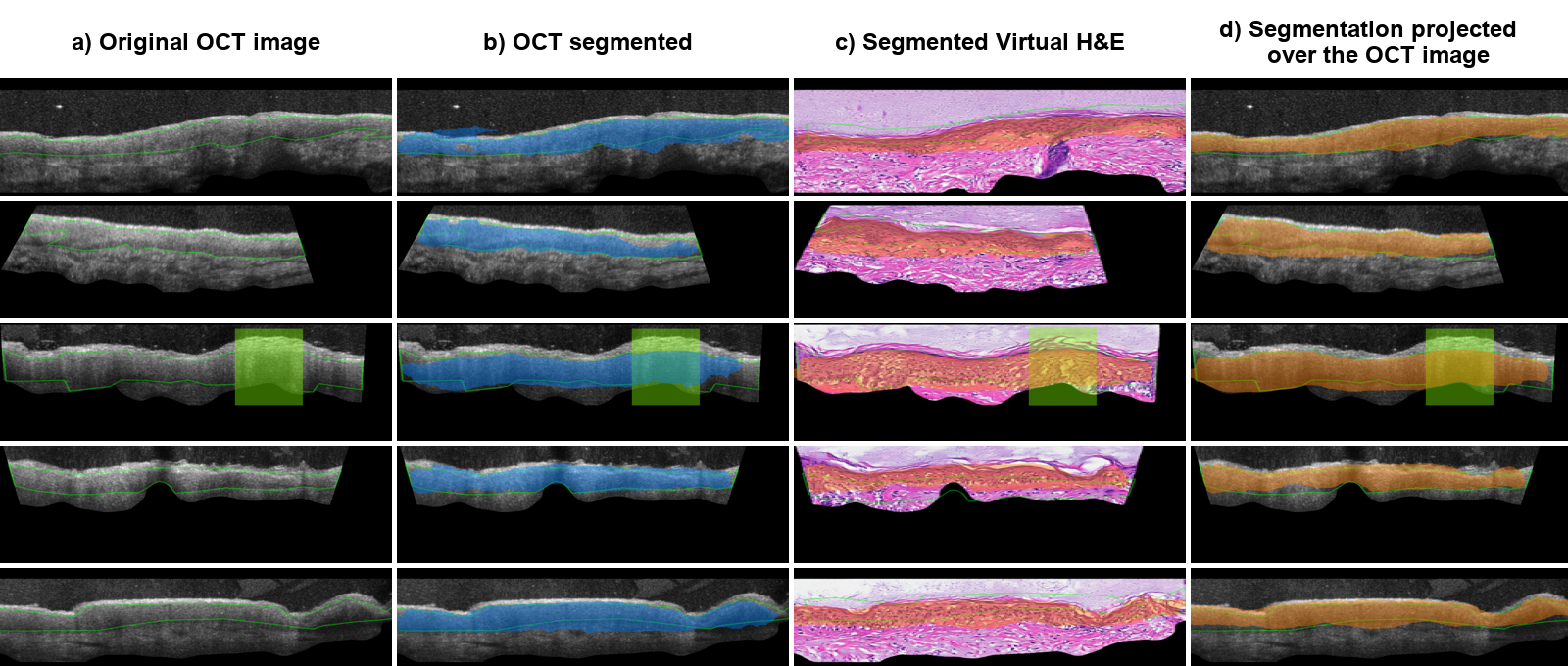}
\caption{\textbf{Segmented areas by MedSAM}. \additionalexamplescaption}
\label{sup:fig:more_segmented_areas_medsam}
\end{figure}

\begin{figure}[H]
\centering
\includegraphics[width=0.8\linewidth]{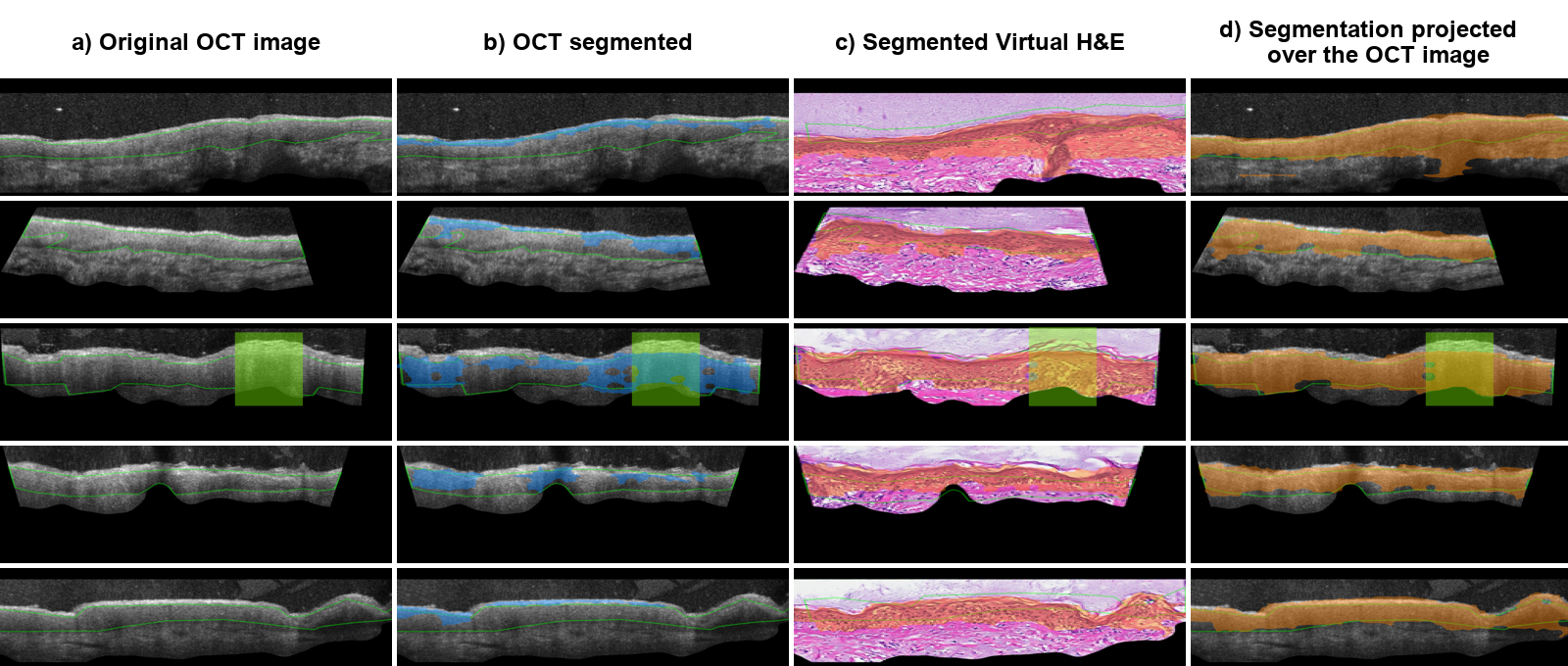}
\caption{\textbf{Segmented areas by SAM-Med2D}. \additionalexamplescaption}
\label{sup:fig:more_segmented_areas_sammed2d}
\end{figure}
}{}

\section*{FoundationShift Applied to Visual Language Foundation Model Classification of OCT Scans}
\begin{figure}[H]
\centering
\includegraphics[width=0.8\linewidth]{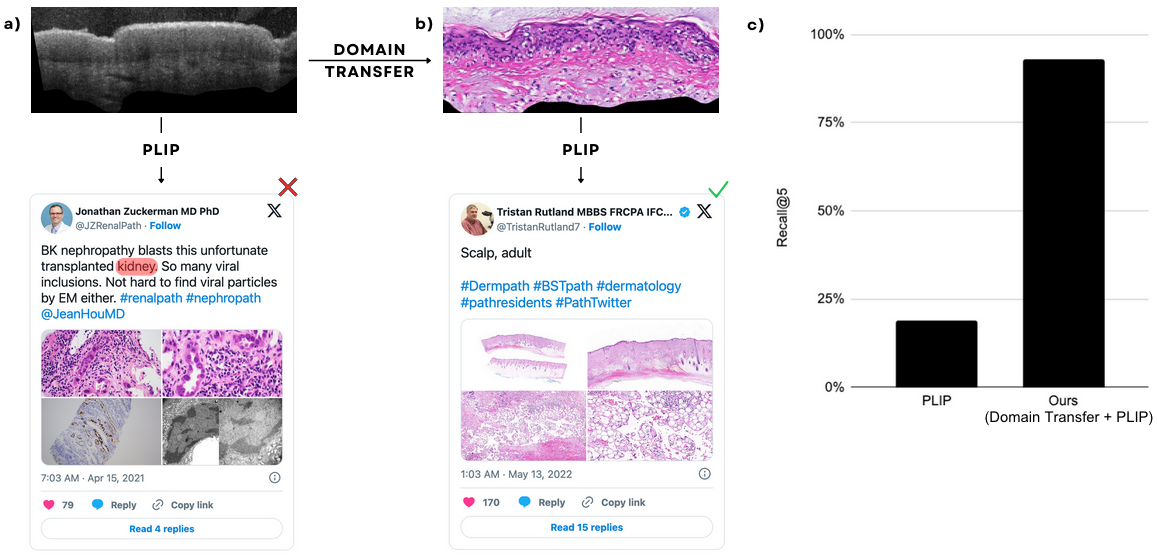}
\caption{\textbf{FoundationShift Applied to Visual Language Foundation Model of OCT Scans. a)}. When evaluating PLIP on an OCT image, PLIP is unable to locate many similar post from Twitter. \textbf{b)} When using domain transfer prior to PLIP, recall increases. \textbf{c)} We evaluated Recall@5 over \numberofsectionsinstudy with and without FoundationShift we were able to increase PLIPS's recall@5 from 19\% (PLIP) to 93\% (FoundationShift + PLIP)
}
\label{sup:fig:oct_skin_plip}
\end{figure}

Huang \etal introduced PLIP~\cite{huang2023visual}, a visual-language foundation model which allows for image classification of pathology H\&E slides. To the best of our knowledge, no visual–language foundation model exist for OCT.
Here we combine domain transfer followed by PLIP to classify OCT images of skin. By utilizing domain transfer, we were able to increase model's recall@5 from 19\% (PLIP) to 93\% (FoundationShift + PLIP) over  \numberofsectionsinstudy images (Figure \ref{sup:fig:oct_skin_plip}). We note that FoundationShift's recall@5 accuracy (93\%) is comparable to recall@5 of ground truth H\&E images in our dataset (82\%). As we noted before~\cite{winetraub2021oct2hist}, virtual H\&E images tend to have less histology processing artifacts compared to ground truth H\&E which may improve recall accuracy.

Similarly to Huang \etal, Recall@5 was computed by evaluating PLIP on images from each modality and examining the tweets that were returned by the model. If a tweet contained dermatology specific description, we considered it as a successful recall. We evaluated PLIP using web interface (https://huggingface.co/spaces/vinid/webplip).

\section*{FoundationShift Applied to Large Language Model Analysis of OCT}
Generative Pre-Trained Transformers such as GPT-4o~\cite{achiam2023gpt} have greatly improved our ability to analyze visual information and provide automated insights. In a recent paper, GPT-4o has been used to analyze and generate histopathology reports over H\&E images. Although GPT-4o seems to have promising performances in analyzing H\&E images. It is currently unable to analyze images of OCT.

\begin{figure}[H]
\centering
\includegraphics[width=0.6\linewidth]{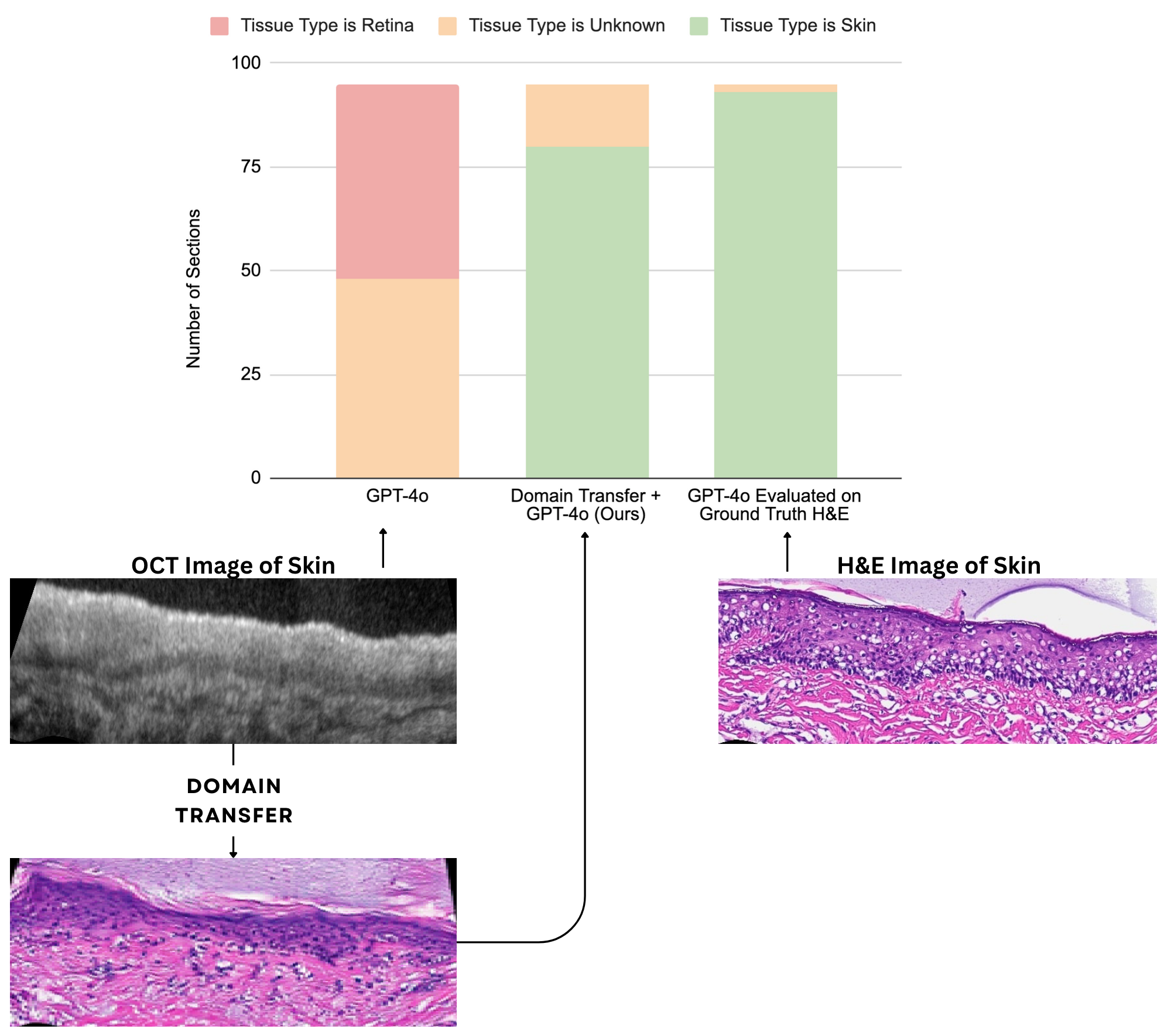}
\caption{\textbf{FoundationShift Applied to Large Language Model Analysis of OCT and H\&E Scans.}
When evaluating GPT-4o on OCT images, 50\% of responses don't specify any tissue region where other 50\% of responses incorrectly identify the region as retina. When utilizing domain transfer first, GPT-4o is able to identify skin as the tissue in more than 84\% of responses. When evaluating GPT-4o on ground truth H\&E images accuracy increases to 98\%. In this analysis we used \numberofsectionsinstudy images in each category. Example responses are provided in Table \ref{sup:oct_chat_gpt_example}.}
\label{sup:fig:oct_skin_chatgpt}
\end{figure}

Utilizing \numberofsectionsinstudy OCT and virtual H\&E images, we asked GPT-4o to analyze each image individually utilizing the promt "Describe the Image" and upload the corresponding image. Example results are found in Table \ref{sup:oct_chat_gpt_example}. GPT-4o is able to correctly report imaging modality (OCT vs. H\&E) at 100\% precision. However, it has difficulty correctly identifying the region where the sample originated from (Figure \ref{sup:fig:oct_skin_chatgpt}). It's not surprising since OCT images of skin are likely a small part of ChatGPT's training set. We were able to significantly increase accuracy of ChatGPT by utilzing domain transfer first.

\begin{table}[H]
    \centering
    \begin{tabular}{|c|c|p{10cm}|}  
        \hline
        Modality & Example Image & Response \\
        \hline
        OCT & \raisebox{-12mm}{\includegraphics[width=4cm]{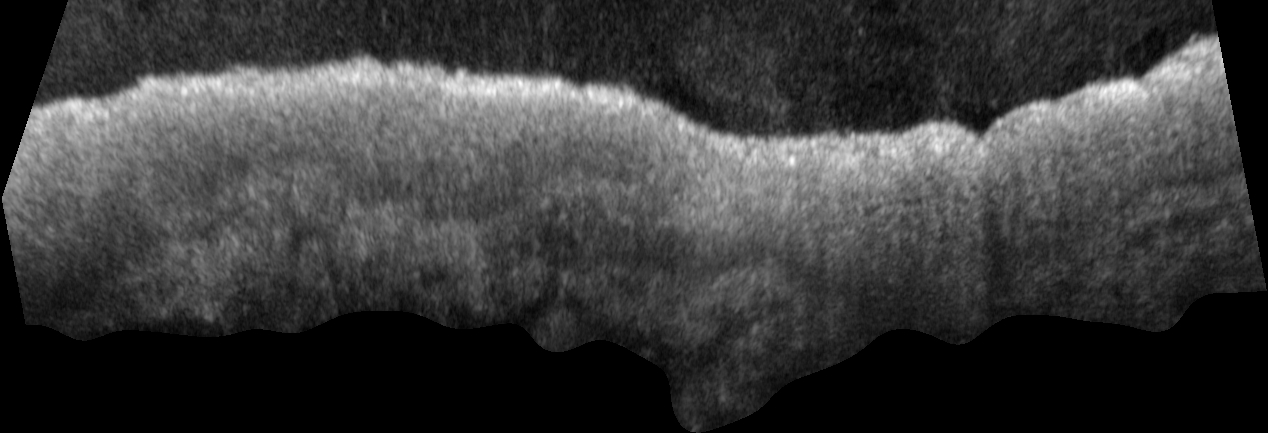}} &         "The image appears to be from an \textbf{OCT device}. In this image: The bright bands and layers correspond to different \textbf{retinal structures}. The undulating dark and light sections suggest changes in the reflective properties of the retinal tissue…" \\
        \hline
        Virtual H\&E & \raisebox{-18mm}{\includegraphics[width=4cm]{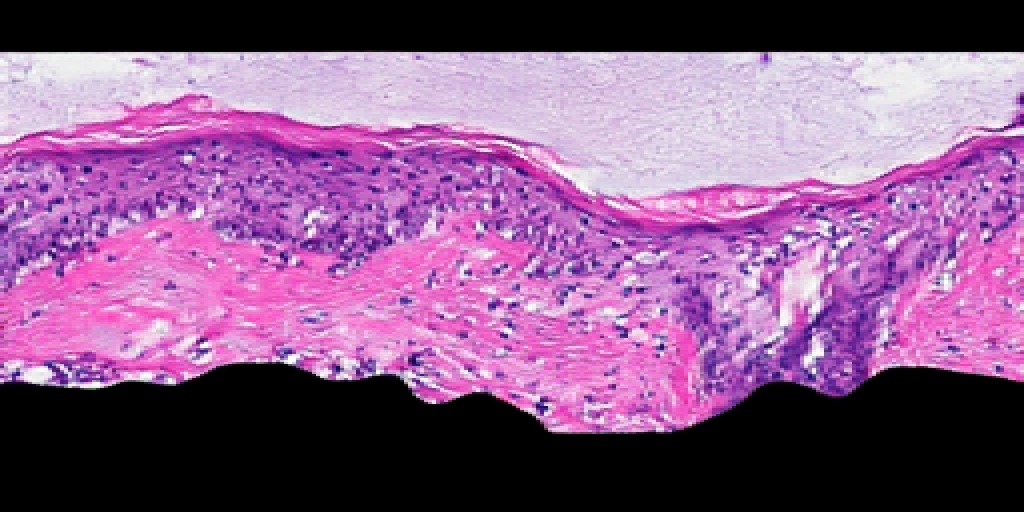}} &         "This image appears to be a histological section of \textbf{skin tissue}… \textbf{Epidermis} (topmost layer): You can observe multiple layers of cells, indicating the different stages of keratinocyte maturation. The outermost layer, which may be the \textbf{stratum corneum}, is visible … \textbf{Dermis}: … It appears wavy and is rich in collagen fibers…" \\
        \hline
    \end{tabular}
    \caption{GPT-4o analysis of example OCT and virtual H\&E examples.}
    \label{sup:oct_chat_gpt_example}
\end{table}

\section*{FoundationShift Applied to Visual Language Foundation Model Classification of RCM}

UNI\cite{chen2024towards} is a general-purpose self-supervised model for computational pathology, pretrained on over 100 million images from more than 100,000 diagnostic H\&E-stained whole-slide images across 20 major tissue types. It demonstrates state-of-the-art performance on various computational pathology tasks and introduces new capabilities such as resolution-agnostic tissue classification and disease subtyping generalization.

\begin{figure}[H]
\centering
\includegraphics[width=0.77\linewidth]{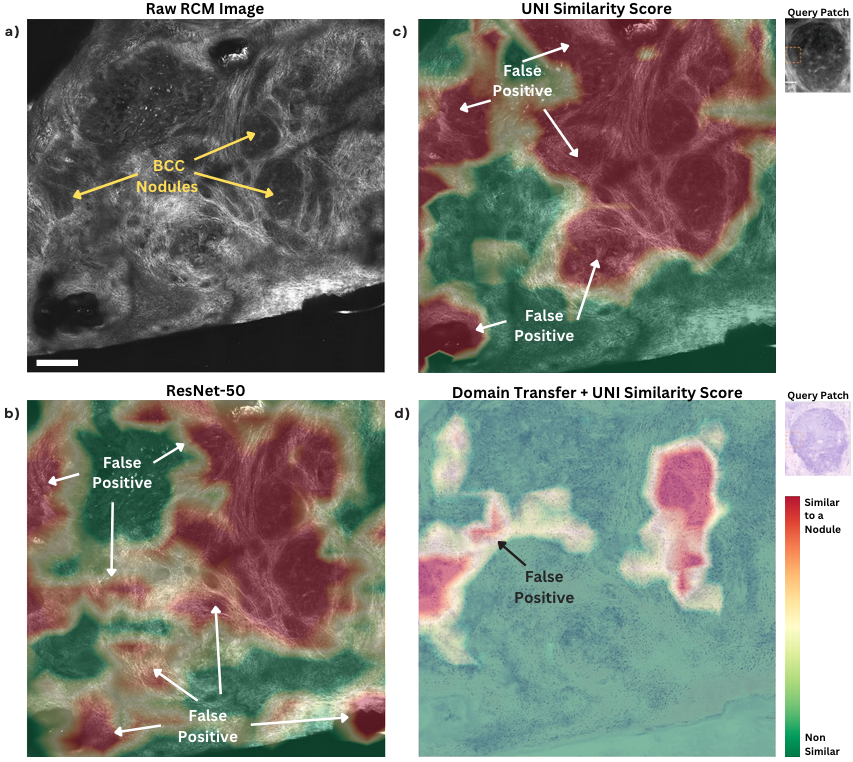}
\caption{
\textbf{FoundationShift Applied to CPath Image Classification Model.}
\textbf{a)}
A mosaic RCM image with BCC nodules was obtained from Li \etal and anotated by our team.
\textbf{b)} ResNet-50 cosine similarity to BCC nodule query patch. 
\textbf{c)} UNI cosine similarity to BCC nodule query patch.
\textbf{d)} Our method (domain transfer + UNI) cosine similarity to BCC nodule query patch.
Scale bar is \SI{200}{\micro\metre}.
}
\label{sup:fig:rcm_skin_uni}
\end{figure}

To be best of our knowledge, no similar model or dataset exists for RCM. We utilized FoundationShift (domain transfer + UNI) to classify BCC nodules in RCM image from a query patch (one shot). In Figure \ref{sup:fig:rcm_skin_uni} we report on qualitative improvement in the ability to identify BCC nodules in an RCM image in one RCM mosaic example provided by Li \etal~\cite{li2021biopsy}.

We evaluated the ability of the ResNet-50~\cite{he2016deep_resnet50}, UNI, and our (domain transfer + UNI) image classifiers to identify BCC nodules in RCM images. To do this, we divided each RCM image into overlapping patches, each measuring 256x256 pixels, with a 64-pixel step overlap. We computed an embedding vector for each patch using each algorithm and measured the cosine distance to the embedding of a query patch. The query patch was a 256x256 image of a BCC nodule from Li \etal Cosine distance was represented by a color gradient ranging from green (low cosine similarity) to red (high cosine similarity), with the mapping chosen to maximize the visibility of BCC nodules in red and other tissue in green. This mapping was selected to minimize false negatives, which are the most problematic for clinical utility.

Qualitatively, we can identify false negative regions in each image and compute F1 score for each algorithm. ResNet-50 presents an F1 score of 0.46, UNI with F1 of 0.35 and our method (domain transfer + UNI) shows an F1 score of 0.85.

\section*{Imaging Setup Modifications Needed for Training a Domain Transfer Network}

To effectively utilize FoundationShift, a domain transfer network is required to convert optical images to H\&E-like images ("virtual staining"). These networks are modality and organ-specific, necessitating a well-curated dataset for training. Several research groups ~\cite{winetraub2021oct2hist,li2021biopsy,kang2022deep,liu2024virtual,shithickv,cao2023label} have developed methodologies to curate training datasets and virtual staining models.

A common prerequisite across these methodologies is the need to co-register the optical imaging modality with an H\&E slide of the exact same region. The accuracy of this registration can vary across different methods. 
Broadly speaking, the most accurate results are achievable when noninvasive images are precisely registered with corresponding histology images to cellular precision, typically within tens of microns.

In this section, we will briefly describe the optical setup modifications performed by Winetraub \etal~\cite{winetraub2021oct2hist} to an OCT system in order to achieve 25-micron accuracy in registration. Winetraub\etal achieve this precise registration by encapsulating a sample in fluorescent gel and photobleaching a pattern onto the gel. The pattern can be detected on a histology slide using a conventional fluorescence microscope prior to H\&E staining.

To photobleach the pattern, a second laser is coupled to the imaging fiber, allowing the system to switch between "imaging mode" and "photobleaching mode" using the same optical setup. It is important to note that the photobleaching laser should have a lower wavelength compared to the imaging laser, to prevent unintended photobleaching of the tissue during imaging.

After repeating the process for 1,000 images, Winetraub\etal utilized an off-the-shelf Pix2Pix network to create a domain transfer model that converts OCT images to H\&E-like images.

\end{document}